\pdfoutput=1

\documentclass[11pt]{article}

\usepackage[preprint]{acl}

\usepackage{times}
\usepackage{latexsym}
\usepackage{textcomp}
\usepackage{microtype}
\usepackage{hyperref}
\usepackage{makecell}
\usepackage{url}
\usepackage{tabularx}
\usepackage{graphicx} %
\usepackage{multirow} %
\usepackage{booktabs} %
\usepackage{diagbox} %
\usepackage{subcaption}
\usepackage{amsmath}
\usepackage{colortbl} %
\usepackage{soul}
\usepackage[framemethod=tikz]{mdframed}
\usepackage{amssymb}

\usepackage[T1]{fontenc}

\usepackage[utf8]{inputenc}

\usepackage{microtype}

\usepackage{inconsolata}

\usepackage{graphicx}

\title{Learning to Plan Long-Term for Language Modeling}

\author{Florian Mai, Nathan Cornille, Marie-Francine Moens \\
  Department of Computer Science \\
  KU Leuven \\
  Leuven, Belgium \\
  \texttt{\{florian.mai, nathan.cornille, sien.moens\}@kuleuven.be}
}

\begin{document}
\maketitle\begin{abstract}
Modern language models predict the next token in the sequence by considering the past text through a powerful function such as attention.
However, language models have no explicit mechanism that allows them to spend computation time for planning long-distance future text, leading to a suboptimal token prediction.
In this paper, we propose a planner that predicts a latent plan for many sentences into the future.
By sampling multiple plans at once, we condition the language model on an accurate approximation of the distribution of text continuations, which leads to better next token prediction accuracy. In effect, this allows trading computation time for prediction accuracy.
\end{abstract}

\section{Introduction}

By pretraining on the next-token prediction objective, autoregressive decoder-only models based on e.g. Transformers attain a variety of skills, spending a small amount of compute for each token.
As such they can be considered \emph{fast, intuitive reasoners}~\citep{DBLP:journals/cacm/BengioLH21}, analogous to the type 1 reasoning systems found in humans according to the dual-process theory~\citep{evans1984heuristic, kahneman2011thinking}. System 1 allows solving intuitive tasks such as perception and talking, but it is insufficient for tasks that require planning, such as writing coherent, long stretches of text.
For planning tasks, humans instead invoke a \emph{slow, deliberate} reasoning system 2.
Most works that attempt to integrate deliberate planning and reasoning ability into LLMs pose the problem as a \emph{post-training} process: by finetuning on reasoning datasets~\citep{hendrycks_measuring_2021, havrilla2024teaching}, by learning to invoke external task-specific planners~\citep{schick2023toolformerlanguage, nye2021improving}, or by employing advanced prompting methods like Chain-of-Thought~\citep{wei_chain--thought_2023}.
However, %
neuroscientific studies have revealed that \emph{predictive coding}, the ability to continuously predict, update and draw on multiple hypotheses about future inputs, is central to language learning and production~\citep{casillas2013development, ylinen2017predictive, shain2020fmri, aitchison2017or, kellogg2013model, mallahi2019role}.
This suggests that the ability to plan originates, at least in part, from learning from unlabeled data and should hence be fostered in LLMs \emph{during pre-training}.
\citet{cornille2024learning} propose a pretraining method in which language modeling is factorized 
into 1) first predicting a high-level latent plan via a separate planner module and 2) then conditioning the language model on generated plans when predicting the next token.
However, their method only predicts a single one-step plan, which predicts merely one sentence ahead. As such, it neither performs long-term planning nor allows to draw on multiple hypotheses through variable compute.

In this paper, we propose an extension of the framework by \citet{cornille2024learning} through two crucial changes (Figure~\ref{fig:overview}):
1) We learn a planner that predicts multiple steps ahead to enable long-term predictive coding. 2) We sample a variable amount of hypotheses from the planner to condition the language model on, allowing to trade off computation time for better prediction accuracy.
Our experimental evaluation demonstrates that both changes contribute to improving the language modeling ability.

\section{Related Work}

\paragraph{Predictive coding}
Multi-step predictions in the form of predictive coding have inspired machine learning algorithms in the past for a long time~\citep{rafols2005using}.
They often serve as an auxiliary loss to produce better representations for a specific downstream task, e.g., document classification~\citep{trinh2018learning}, POS tagging~\citep{lan2021predictive} and sentence representation learning~\citep{araujo2021augmenting, araujo2023learning}.
For language modeling, \citet{gloeckle2024better} recently extended the next-token prediction objective to predicting $n$ tokens ahead. They observe that multi-step prediction yields up to 17\% better performance on coding tasks, demonstrating the potential of multi-step prediction for reasoning tasks. However, the improvement only appears with large-scale training.
In our work, multi-step predictions are not an auxiliary task, but \emph{directly} inform the downstream language model in its prediction.

\paragraph{Additional inference-time compute}
Aiming to overcome the computational limitations of the original Transformer architecture, \citet{dehghani2018universal} equip it with Adaptive Computation Time~\citep{graves2016adaptive}.
Many works attempt to transfer AlphaGo's famous success in Go~\citep{silver_mastering_2017} to text by generating and evaluating multiple paths of \emph{concrete} text to improve performance~\citep{yao_tree_2023-1, wang_self-consistency_2023, zelikman2024quiet}. %
In contrast, our approach generates paths in an \emph{abstract} space, which is more akin to MuZero~\citep{schrittwieser_mastering_2019}.

\section{Methods}

\begin{figure}[t]
    \centering
    \includegraphics[width=\columnwidth]{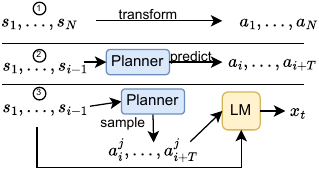}
    \caption{Overview of our method.}
    \label{fig:overview}
\end{figure}
The key idea of the method is to transform an unlabeled text corpus into sequences of abstract writing actions and use these actions to guide the language model.
Our method consists of three steps (cmp. Figure~\ref{fig:overview}):
\textcircled{1}: Inferring action sequences from unlabeled texts \textcircled{2}: Training a multi-step planner to predict the next actions \textcircled{3}: Sampling multiple paths from the planner to condition the LM.

\subsection{Training an External Planner}\label{sec:planner-framework}
We briefly review the method of \citet{cornille2024learning}, who train a planner that can predict only one step into the future.
In Section~\ref{sec:multi-step-planning}, we propose a novel multi-step planner. In Section~\ref{sec:multi-sample-conditioning}, we propose a novel way of conditioning the LM on multiple sampled action sequences. 

Given a training corpus $X$ with articles $X = t_1, t_2, \ldots, t_n$, we first embed each text unit $t_i$ into a low-dimensional vector $\mathbf{z}_i = \text{E}(t_i)$ using a text encoder $\text{E}$.
We then cluster these embeddings into $C$ clusters via k-means.
Since the cluster centroids do not represent concrete sentences, \citet{cornille2024learning} call them "abstract writing actions"  $a \in \mathcal{A}$.
This labeling process transforms the article $X = t_1, t_2, \ldots, t_n$ into $X' = a_1, t_1, a_2, t_2, \ldots, a_n, t_n$.

The planner module $\text{P}$ is composed of two functions: the representation function $h$ and the prediction function $f$.
The function $h$ turns the textual context $t_1, t_2, \ldots, t_{i-1}$ into a set of latent variables $\mathbf{z}_1, \mathbf{z}_2, \ldots, \mathbf{z}_{i-1}$ by using a text encoder $\text{E}$ per sentence:
\begin{align*}
\mathbf{z}_{1:i-1} & = h(t_1, t_2, \ldots, t_{i-1}) \\ & = \{\text{E}(t_1), \text{E}(t_2), \ldots, \text{E}(t_{i-1})\}
\end{align*}
The function $f$ consists of a Transformer encoder that first contextualizes $\mathbf{z}_1, \mathbf{z}_2, \ldots, \mathbf{z}_{i-1}$, averages them after the last layer, and finally passes the resulting vector into a linear classifier to return a probability distribution over the possible writing actions.
The predicted action results as:
\begin{equation}
\hat{a}_i = \operatorname{argmax}\limits_{a \in A} f(\mathbf{z}_{1:i-1})
\end{equation}
During training of the language model, at every sentence boundary the planner module $\text{P}$ predicts the next writing action $\hat{\mathbf{a}}_{i}$ based on the current context $t_1, t_2, \ldots, t_{i-1}$.
The language model $\text{LM}$ is then conditioned on $\hat{a}_{i}$ when generating the next sentence $t_{i}$.
The training objective is to predict the next token based on the previous words and the predicted actions, approximating the distribution $p(x_t | x_{1:t-1}, \hat{a}_{1:i})$.

A simple adapter module integrates the action information into the language model through a linear projection, $\mathbf{c}_i = \mathbf{W}\operatorname{Emb}(\hat{a}_{i}) + \mathbf{b}$.
Finally, $\mathbf{c}_i$ is added to every token embedding in sentence $t_i$.

\subsection{Planning Multiple Steps Ahead}\label{sec:multi-step-planning}

To enhance the planner's capability, we extend it to predict multiple steps into the future.
Instead of predicting only the next action $\hat{a}_{i}$, the planner now generates a sequence of future actions $\hat{a}_{i}, \hat{a}_{i+1}, \ldots, \hat{a}_{i+T}$, where $T$ represents the number of future timesteps considered.

While the representation function $h$ and prediction function $f$ remain, we introduce an additional dynamics function $g$, which transforms the latent representation depending on the predicted action.
Hence, for the $a_{i+k}$-th writing action ($0 \leq k < T$), we first apply $g$ recurrently on the output of $h$ $k$ times, and then use $f$ to predict the next action:
\begin{align*}
\mathbf{z}_{1:i-1} &= h(t_1, t_2, \ldots, t_{i-1}) \\
\mathbf{z}_{1:i+k} &= g(\mathbf{z}_{1:i+k-1}, \hat{a}_{i+k-1}) & \forall 1 \leq k \leq T - 1 \\
\hat{a}_{i+k} &= f(\mathbf{z}_{1:i+k}) & \forall 0 \leq k \leq T - 1
\end{align*}

The function $g$ works as follows: 
First, we plug $\hat{a}_{i+k-1}$ into an action embedding table $\operatorname{Emb}$ and add it to the set of $\mathbf{z}_1, \ldots, \mathbf{z}_{i+k-1}$.
Then, we use a transformer encoder on top of it, which gives us our new hidden state $\mathbf{z}'_1, \ldots, \mathbf{z}'_{i+k}$.
Formally:
\begin{align*}
& g(\mathbf{z}'_{1:i+k-1}, \hat{a}_{i+k-1}) \\
= & \text{Transformer}(\mathbf{z}_1, \ldots, \mathbf{z}_{i-1}, \operatorname{Emb}(\hat{a}_{i+k-1}))
\end{align*}

Note that our multi-step planner factorizes 
\begin{align*}
    & p(a_{i+1} \ldots a_{i+T} | t_1 \ldots t_i) \\
     =  & \prod_{j=1}^T p(a_{i+j} | a_{i+1}\ldots a_{i+j-1},t_1 \ldots t_i),
\end{align*}
allowing for efficiently sampling action sequences autoregressively.

\subsection{Multi-path Adapter}\label{sec:multi-sample-conditioning}

During inference at text unit $i - 1$, instead of using the single best action (argmax) $\hat{a}_{i}$ of the first step, we sample $K$ paths $\hat{a}^{j}_{i:i+T}, 1 \leq j \leq K$ from the planner with temperature $\tau$\footnote{$\tau = 1.0$ is a reasonable default choice, see App.~\ref{app:softmax-temperature}.}.
These $K$ paths allow the language model to account for a diverse set of possible futures, enhancing its ability to generate coherent long-term text.

A straight-forward adaptation of the adapter module by \citet{cornille2024learning} can be constructed as follows: For each path $j$, we simply average the linearly projected action embeddings $\mathbf{\tilde{c}}_i^j = \frac{1}{T} \sum_{t = 0}^{T-1}\mathbf{c}^{j}_{i + t}$ to obtain a representation of the path.
Then, a final representation $\mathbf{\hat{c}}_i$ is obtained as $\mathbf{\hat{c}}_i = \frac{1}{K} \sum_{j = 1}^{K}\mathbf{\tilde{c}}^j_{i}$.
In the experiment section, we refer to this as \emph{Project and Avg}.
However, this adaptation has several shortcomings: 1) It completely disregards the sequential structure of actions in a path, and 2) it is unable to compute nonlinear interactions between multiple paths.

To enable the language model to effectively reason over multiple paths, we introduce a new adapter architecture consisting of a \emph{PathTransformer} ($\operatorname{PT}$), which is responsible for aggregating a single path into a vector that represents the path, and a \emph{SampleTransformer} ($\operatorname{ST}$), which aggregates a set of path vectors.
Both models are bi-directional encoder-only transformers~\citep{vaswani_attention_2017}. 
For enabling better training stability, we additionally found it necessary to apply a ReZero-inspired~\citep{bachlechner2021rezero} normalization which initializes the solution close to the naive adapter. Formally, the model is described as:
\begin{align}
\mathbf{c}^j_i &= \operatorname{PT}(\hat{\operatorname{Emb}(a)}_{i:i+T}^j + \mathbf{p}_{1:T}) \cdot \alpha_1 + \mathbf{\tilde{c}}_i^j\\
 \mathbf{c}_i &= \operatorname
{ST}(\mathbf{c}_i^1, \mathbf{c}_i^2, \ldots, \mathbf{c}_i^K) \cdot \alpha_2 + \mathbf{\hat{c}}_i
\end{align}
$\mathbf{p}_{1:T}$ are absolute position embeddings indicating the order of actions in a path. $\alpha_1, \alpha_2 \in \mathbb{R}$ are learnable scalars initialized to zero.

\section{Experiments}
The purpose of our experiments is to demonstrate the benefit of our contributions for language modeling: 1) Multi-step planning and 2) conditioning on multiple sampled plans.

\paragraph{Baselines and metrics} \citet{cornille2024learning} can be viewed as a special case of our model with $T = 1, K = 1$ and \emph{Project and Avg} adapter. It thus serves as our primary baseline.
Furthermore, we reproduce the \textbf{Fixed} baseline from \citet{cornille2024learning}, which reports the LM performance with finetuning with a single, fixed action only, demonstrating the usefulness of conditioning on planner-generated outputs.
As is standard practice for language modeling, all models are evaluated via the perplexity metric.
For reference only, we report the edit distance metric proposed by \citet{cornille2024learning}, which indicates how well generated text follows the ground truth in terms of action sequences.

\paragraph{Hyperparameters} Following \citet{cornille2024learning}, all experiments are performed based on GPT-2 small (128M parameters) finetuned on 285310 articles of English Wikipedia. The full set of hyperparameters is reported in Appendix~\ref{sec:appendix:hyperparams}.

\subsection{Results}
Table~\ref{tab:main_results} shows the results of our model with various configurations in comparison to the baselines. All models are tested with the same $K$ as in training. 

\paragraph{Impact of multi-step predictions}
Considering a fixed amount of path samples $K = 10$, when moving from $T = 1$ to $T = 5$, the perplexity of our model improves substantially by 0.2.
When moving from $T = 5$ to $T = 10$, the improvement continues albeit relatively small.
We attribute this to high uncertainty when modeling long-distance futures.

\begin{figure}[t]
    \centering
    \includegraphics[width=\columnwidth]{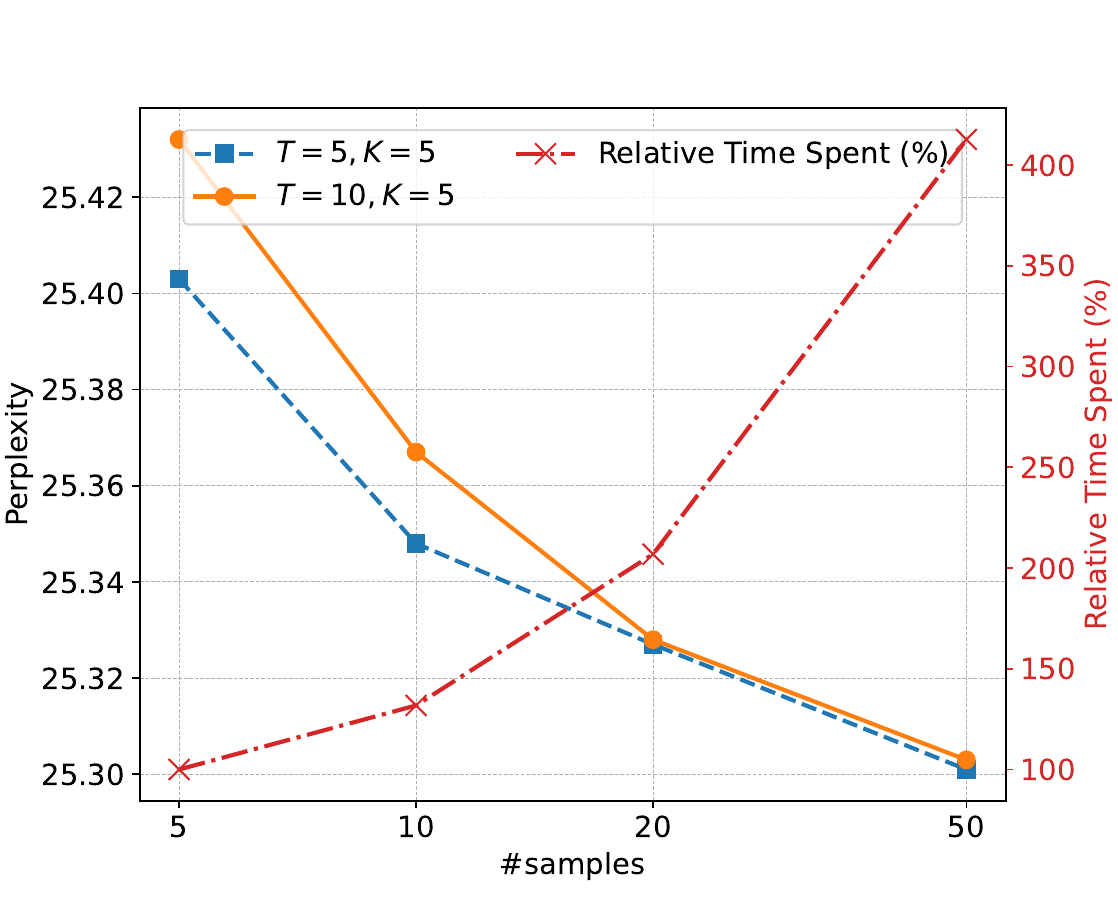}
    \caption{Performance and relative generation time as a function of the number of samples $K$ drawn.}
    \label{fig:num-samples-generalization}
\end{figure}

\paragraph{Impact of conditioning on multiple paths}
Considering a fixed number of time steps $T$,
the performance of our model also improves consistently when conditioned on an increasing number $K$ of sampled paths (Table~\ref{tab:main_results}).
In order to understand whether this generalizes to larger $K$ than seen during training, in Figure~\ref{fig:num-samples-generalization} we increase the number of sampled paths $K$ \emph{at inference time only}.
This experiment demonstrates that the performance continues to improve until at least $K=50$.
Naturally, this comes at the expense of additional compute.

While our best models clearly outperform \citet{cornille2024learning}, for $K = 1$, our model performs worse. We explain this with the fact that sampling once is generally worse than argmax.

\begin{table}[h]
\centering
\begin{tabular}{|l|c|c|c|}
\hline
\textbf{Model} & \textbf{PPL} (\textdownarrow) & \textbf{Edit} (\textdownarrow) \\ \hline
\multicolumn{3}{|c|}{\textbf{Baselines}} \\ \hline
Fixed & 26.67 & 4.67 \\ \hline
\citet{cornille2024learning} & 25.54 & 4.48  \\ \hline
\hline
\multicolumn{3}{|c|}{\textbf{Ours (T = 1)}} \\ \hline
K = 10  & 25.56 & 4.53  \\ \hline
\hline
\multicolumn{3}{|c|}{\textbf{Ours (T = 5)}} \\ \hline
K = 1 & 25.88 & 4.47  \\ \hline
K = 5  & 25.44 & 4.52  \\ \hline
K = 10 & 25.35 & 4.47  \\ \hline
\hline
\multicolumn{3}{|c|}{\textbf{Ours (T = 10)}} \\ \hline
K = 1 & 25.76 & 4.51  \\ \hline
K = 5 & 25.40 & 4.41  \\ \hline
K = 10 & 25.32 & 4.45  \\ \hline
\hline
\multicolumn{3}{|c|}{\textbf{Ablations (T = 5, K = 10)}} \\ 
\hline
 Full model & 25.35 & 4.47  \\
\hline
 Project and Avg & 25.49 & 4.44 \\ \hline
 No ReZero connection & 25.78 & 4.30 \\
 \hline
\end{tabular}
\caption{All results are based on GPT-2. PPL represents perplexity, and Edit represents the edit distance.}
\label{tab:main_results}
\end{table}

\paragraph{Ablations}
In Table~\ref{tab:main_results} (bottom), we measure the effect of our proposed multi-path adapter module (cmp. Section~\ref{sec:multi-sample-conditioning}). 
First, using the naive \textbf{Project and Avg} adapter instead of our proposed Sample- and PathTransformers performs worse by 0.14 PPL.
Second, the \textbf{No ReZero connection} ablation increasing PPL by 0.43 shows the importance of initializing the solution of the adapter close to the \textbf{Project and Avg} model to enable proper learning.

\subsection{Discussion}
Our consistent improvements in perplexity indicate that both integrating long-term predictions of the future writing process and modeling multiple future paths provide an LM with information that is valuable even for making local predictions.
Consequently, our model outperforms the single-step planner by \citet{cornille2024learning}.

Moreover, a core motivation of our work is to allow a language model to spend additional test-time compute to improve its predictions, similar to how AlphaGo~\citep{silver_mastering_2017} uses a lot of inference-time compute to achieve superhuman performance in Go. Demonstrating that our model, too, can trade off compute for better performance, we take a first step towards enabling this property for LMs.

\section{Conclusion}
LLMs acquire many skills through the next-token prediction objective, but planning remains a major weakness.
We take a step towards learning to plan from pretraining on unlabeled data by predicting long sequences of abstract writing actions. By allowing the LM to condition on an arbitrary amount of sampled sequences, our model can flexibly trade off compute for prediction accuracy.
This opens exciting research directions for planning with LMs.

\newpage

\subsubsection*{Acknowledgments}
This research was financed by the CALCULUS project—Commonsense and Anticipation enriched Learning of Continuous representations—European Research Council Advanced Grant H2020-ERC-2017-ADG 788506, \url{http://calculus-project.eu/}.

\section*{Limitations}

\paragraph{Lack of large-scale experiments}
Our work is motivated by the promise of integrating a slow, deliberate reasoning system into the framework of standard language models.
We validate our proposed approach through controlled experiments that require the training of many models.
Therefore, the evaluation in this paper is limited to the relatively small language model GPT2-small with 128M parameters.
However, we have two reasons to believe that our approach will generalize to larger scale as well.
First, \citet{cornille2024learning}, who propose the framework on which we build, show that the framework yields improvements for the relatively large LLM OLMo-1B~\cite{groeneveld2024olmo} as well.
Second, \citet{gloeckle2024better} recently showed that their proposed pretraining objective, which, like ours, predicts multiple steps ahead, shows even greater potential at large model sizes starting from 7B.
Since our related approach already shows promising results at small scale, we expect it to yield even better performance at larger scale.

\paragraph{Flexibility of compute-performance tradeoff}
Inspired by AlphaGo's success, our method is able to trade off inference-time compute for better next-token prediction accuracy.
However, this can be quite expensive, especially if the maximum amount of compute is spent every time the planner is called, i.e., at every sentence boundary, limiting the practicality of our method in its current state.
To address this limitation in the future, we envision a mechanism similar to Adaptive Computation Time~\citep{graves2016adaptive} that can learn how much additional compute is needed at any point.
Given the success of Universal Transformers~\citep{dehghani2018universal} at incorporating this mechanism, we are confident that this limitation will be resolved in the future.

\paragraph{Edit distance results}
The purpose of our work is to improve language modeling. As the number of time steps $T$ and the number of drawn samples $K$ are increased, our proposed method consistently improves performance in terms of perplexity, the standard metric for language modeling.
However, the performance in terms of edit distance shows no clear trend in either direction.
This indicates that the edit distance, proposed by \citet{cornille2024learning} to measure how well the model generates articles that adhere to the structure of the ground truth article, is noisy. In fact, some of the edit distance results reported by \citet{cornille2024learning} are also in disagreement with the perplexity improvements.
Future work interested in measuring the quality of generations in terms of structure should reconsider this choice of metric.

\section*{Ethical and Broader Impact}
Our paper is concerned with LMs in general, which can be used to generate data that is within the training distribution. We train our models exclusively on Wikipedia, which is a corpus that contains very little to no content that is directly harmful to the user (e.g. slurs, insults, etc.).
However, our developed method can, in principle, be used to enhance any LM, including those trained on harmful data, which is outside our control.

In the past, ethical concerns about LLMs have been raised because they are compute-intensive, energy-intensive, and carbon-intensive~\citep{strubell2019energy, bender2021dangers}. Our paper proposes a method that can trade off more compute for better performance, potentially adding to this problem.
Therefore, rather than increasing the compute indiscriminately, we advocate for researching methods that can \emph{learn} when it is necessary to spend more compute.
We suspect that this avenue will ultimately lead to more energy-efficient LLMs.

\bibliography{manual}

\newpage

\appendix

\section{Impact of Softmax Temperature}\label{app:softmax-temperature}
\begin{figure}[ht]
    \centering
    \includegraphics[width=0.48\textwidth]{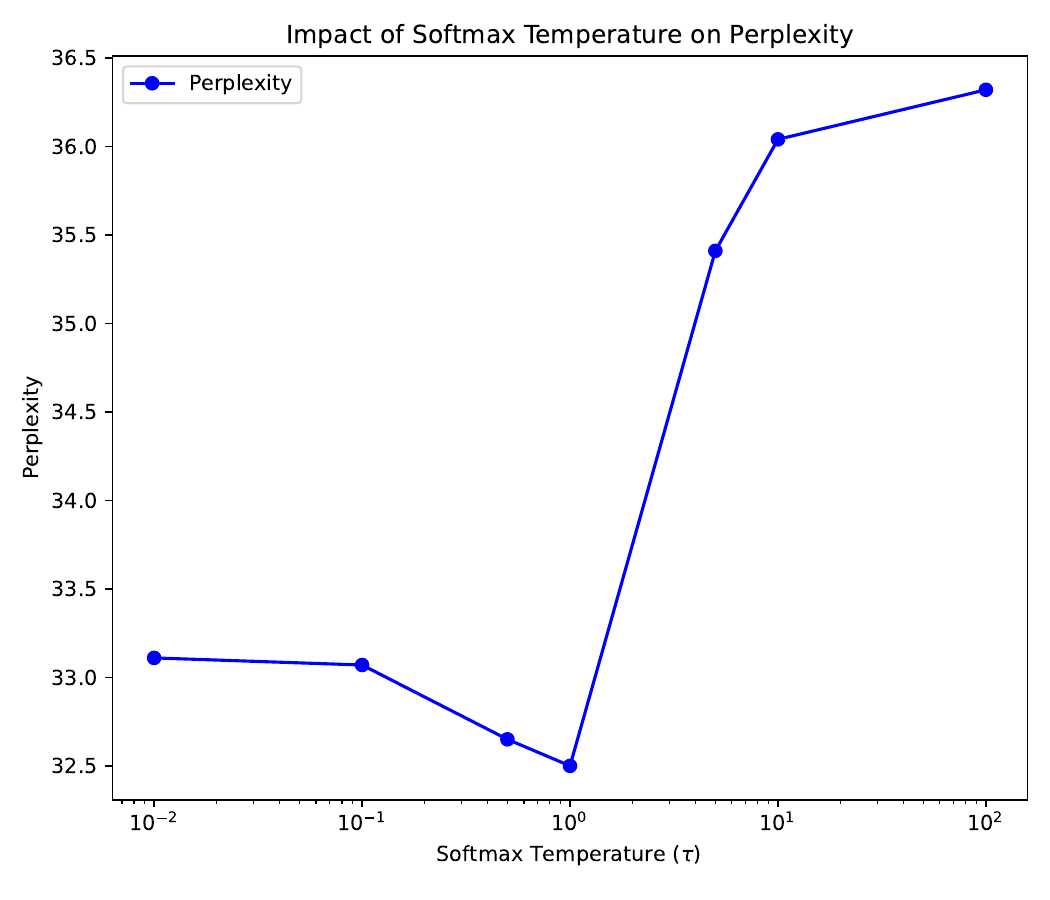}
    \caption{Perplexity on the validation set depending on the sampling temperature $\tau$. Since the textual context in the evaluation on the validation set is shorter, reported perplexities are larger than on the test set.}
    \label{fig:diversity}
\end{figure}

In preliminary experiments, we performed a small experiments to test the impact of the softmax temperature $\tau$ that is applied when sampling action paths.
As Figure~\ref{fig:diversity} shows, $\tau = 1.0$ leads to the lowest perplexity.
When the temperature is too low, the performance degrades because the diversity of sampled paths goes down, decreasing the amount of effective information passed to the LM. When the temperature is too high, the effective probability distribution converges towards uniform, which means that only uninformative paths are passed to the LM.

\section{Implementation Details}
\label{sec:appendix:hyperparams}

\subsection{Implementation}
Our implementation extends upon the source code of \citet{cornille2024learning}, which was privately shared with us.
Once their code is shared publicly, we will release our own extensions as soon as possible thereafter. Unless specified explicitly, all packages use default parameters.

The code base makes use of PyTorch~\citep{paszke2019pytorch}, the Huggingface `datasets`~\citep{lhoest-etal-2021-datasets} and `transformers`~\citep{wolf2020transformers} libraries to load and preprocess data and pretrained models (GPT-2~\citep{radford2019language}), respectively. Furthermore, we used PyTorch-Lightning~\citep{falcon2020pytorchlightning} for model training.

We obtain the Wikipedia dataset through the `datasets` library at \url{https://huggingface.co/datasets/wikipedia} (version `20220301` from March 2022). No additional preprocessing is applied. We randomly subsample 285,310 articles for training, and 1,000 for each validation and test set, respectively. 

Abstract writing actions are generated by first splitting every article into sentences using spaCy~\citep{honnibal2020spacy}, and then encoding them into embeddings using MPNet-base-v2~\citep{song_mpnet_2020} via the SentenceTransformer library~\citep{reimers2019sentence}\footnote{\url{https://huggingface.co/sentence-transformers/all-mpnet-base-v2}} to encode sentences into embeddings.
The final clustering step is performed via Scikit-Learn~\citep{scikit-learn} with k-means++ initialization~\citep{arthur2007k}.
All used libraries are either open source or freely usable for academic purposes.

We ran our experiments on a compute grid with NVIDIA P100s (16GB). Only one GPU was used per experiment.
Pretraining the planner took at max 48h, with the maximum reached when $T = 10$, i.e., necessary compute increases the more steps we predict ahead.
Finetuning the LM takes another 24 hours. This includes first using the planner to predict writing actions for all data in the training set, and then finetuning the LM conditioned on the actions.
Evaluating perplexity takes in total about $1h + 0.25h \cdot K$, while evaluating edit-distance (which requires generation) takes around 3 times as long. 

Most preliminary experiments were ran on a 10$\times$ smaller subset of the data, of which we ran roughly 200 experiments (using 10$\times$ less compute).
Every final experiment was run once. We estimate that in total we used around 4000 GPU hours.

\subsection{Hyperparameters}

\begin{table}[h]
\centering
\begin{tabular}{@{}l|c@{}}
\toprule
Model                         & Parameter Count   \\ \midrule
\hline
Planner  & 122.68M  \\
\hline
Representation function $h$ & 110M\\
Dynamics function $g$ & 6.3M \\
Prediction function $f$ & 6.3M \\
\hline
Language Model & 240.62M  \\
GPT2-Small                    & 124,44M \\
Extra conditioning parameters & 116.18M        \\
\bottomrule
\end{tabular}
\caption{Parameter counts for our models}
\label{tab:model_details}
\end{table}

Table~\ref{tab:model_details} shows the number of parameters used in each model component. Table~\ref{tab:hyperparameters} shows other hyperparameters used for our experiments. We will make the code available upon acceptance.

\begin{table*}[t] %
\centering
\caption{Hyperparameter settings} 
\label{tab:hyperparameters} 
\begin{tabular}{ll} 
\toprule
\textbf{Hyperparameter} & \textbf{Value} \\
\midrule
Context window size      & 128 \\
Train $\mid$ test $\mid$ val split sizes & 285310 $\mid$ 1000 $\mid$ 1000 \\
K-means initialization & k-means++ \\
Number of tokens generated for edit distance & 128 \\
Default action count & 1024 \\
Action embedding dimension & 768 \\
\midrule

\multicolumn{2}{l}{\textbf{Language Model Fine-tuning}} \\ 
Batch size & 32 \\
Learning rate & 1e-4 \\ 
$\tau$ & 1.0 \\ 
\midrule

\multicolumn{2}{l}{\textbf{Planner Training}} \\ 
Batch size & 512 \\
Learning rate & 1e-4 \\ 
\bottomrule
\end{tabular}
\end{table*}

\end{document}